\documentclass{article}

\usepackage[final]{colm2026_conference}

\usepackage{microtype}
\usepackage{hyperref}
\usepackage{url}
\usepackage{booktabs}
\usepackage{amsfonts}
\usepackage{amsmath}
\usepackage{xcolor}
\usepackage{graphicx}
\usepackage{enumitem}
\usepackage{xspace}
\usepackage{tikz}
\usetikzlibrary{arrows.meta,positioning,shapes.geometric,fit,backgrounds,calc}
\usepackage[capitalise,noabbrev]{cleveref}

\usepackage{lineno}
\ifcolmsubmission
\linenumbers
\fi

\definecolor{darkblue}{rgb}{0, 0, 0.5}
\hypersetup{colorlinks=true, citecolor=darkblue, linkcolor=darkblue, urlcolor=darkblue}

\newcommand{\ResultCMSE}{0.0222}
\newcommand{\ResultCMSESE}{0.0108}

\newcommand{\ResultSignOverlap}{0.70}
\newcommand{\ResultSignOverlapSE}{0.03}

\newcommand{\ResultSignOverlapBC}{0.80}
\newcommand{\ResultSignOverlapBCSE}{0.03}

\newcommand{\ResultLaunchAlign}{0.41}
\newcommand{\ResultLaunchAlignSE}{0.06}

\newcommand{\ResultMAE}{0.0893}
\newcommand{\ResultMAESE}{0.0178}

\newcommand{\ResultSignAccuracy}{0.70}
\newcommand{\ResultSignAccuracySE}{0.06}

\newcommand{\ResultNExperiments}{67}

\newcommand{\ResultNaiveCoinFlipSignOv}{0.75}

\newcommand{\ResultNaiveUnifDrawSignOv}{0.66}

\newcommand{\ResultNaiveAlwaysPosSignAcc}{0.57}

\newcommand{\ResultNaiveCoinFlipSignAcc}{0.50}

\newcommand{\ResultSignReplication}{0.69}

\newcommand{\ATE}{\mathrm{ATE}}
\newcommand{\hatATE}{\widehat{\ATE}}
\newcommand{\agentATE}{\widetilde{\ATE}}
\newcommand{\agentHatATE}{\widehat{\widetilde{\ATE}}}

\newtheorem{assumption}{Assumption}

\title{%
    Can AI Agents Simulate A/B Test Outcomes?
    A Validation Framework for Agentic Experimentation%
}

\author{%
  Stefan Hut\thanks{Equal contribution.} \\
  \texttt{hutstefa@amazon.com} \And
  Lorenzo Masoero\footnotemark[1] \\
  \texttt{masoerl@amazon.com}%
}

\begin{document}

\maketitle

\begin{abstract}

A/B testing remains the standard for rolling out new features in the
technology industry.
Each experiment, however, consumes real traffic, engineering effort,
and weeks of wall-clock time.
Can AI agents---conditioned on behavioral profiles and contextual
descriptions of the intervention---simulate outcomes accurately
enough to vet candidate treatments before committing live traffic?
We formalize this question as a \emph{Simulated Randomized Controlled
Trial} (S-RCT) and derive a two-layer error decomposition that
separates agent approximation error from subsampling error, enabling
targeted improvements to each.
The framework is agent-agnostic: any behavioral model---from a
fine-tuned specialist to a general-purpose foundation model---can
serve as the simulation engine.
Validated on \ResultNExperiments{} historical marketing A/B tests, a
baseline S-RCT using an off-the-shelf foundation model systematically
overshoots effect magnitudes.
A two-phase pre-period calibration protocol corrects much of this
magnitude error, reducing the squared prediction error (after removing
irreducible measurement noise) by ${\sim}77\times$; a within-subject
design---where each agent is exposed to both arms---further reduces
standard errors by ${\sim}2.4\times$.
On naive directional metrics the simulator looks promising, agreeing
with the historical sign for roughly $70\%$ of tests (sign accuracy
$\ResultSignAccuracy$, sign overlap $\ResultSignOverlap$); but measured
against simple uninformed baselines and the sign-replication ceiling of
the A/B tests, this agreement is not clear evidence of directional
signal on this noisy set.
We discuss limitations of the current approach and identify
applications where experimenters stand to benefit from agentic
signals.

\end{abstract}


\section{Introduction}
\label{sec:intro}

Online controlled experimentation (A/B testing) is how technology
services learn what works for customers, enabling safe, systematic
iteration over product experiences at
scale~\cite{kohavi2009controlled,kohavi2020trustworthy}.
Yet each experiment requires engineering integration, weeks of
wall-clock time to accumulate statistical power, and exposure of real
customer traffic to the intervention under study.
At the scale of a large organization running thousands of
experiments per year, the aggregate cost is substantial: each
experiment consumes engineering time, traffic allocation, and analysis
overhead, while exposing customers to potentially underperforming
treatments.
Teams that iterate through dozens of candidate treatments before
finding a winner bear this cost repeatedly.
This affects applied scientists, product managers, and builders alike.

A sufficiently accurate behavioral simulator could fundamentally
change this dynamic: letting teams vet dozens of candidates cheaply
and reserve live traffic for the most promising few, so that
customers benefit from improvements sooner.
The advent of large foundation models offers a path toward such a
simulator: conditioning a general-purpose model on a behavioral
profile and eliciting decisions directly, without the domain-specific
engineering bottleneck of hand-crafted agent-based
models~\cite{epstein1996agent,tesfatsion2006handbook}.
Importantly, the simulation engine is a pluggable module: while we
instantiate it here with an off-the-shelf foundation model for rapid
iteration, the framework accommodates any behavioral model, including
fine-tuned specialists and non-language-model architectures.
We expect domain-specific models to eventually
outperform general-purpose ones for this task.

The idea of using language models as simulated economic agents dates
to \citet{horton2023large}; recent work on agentic A/B
testing~\cite{wang2025paars,castelo2026simgym,shopr12025} takes the
next step, proposing to simulate user behavior with AI agents and
estimate treatment effects from the simulated outcomes.
Early validation found promising aggregate alignment with human
baselines but systematic individual-level
divergences~\cite{maier2025simulating,anthis2025validity,humanstudybench2026};
subsequent work showed that structured persona pipelines and rich
contextual grounding narrow the
gap~\cite{persona2026,contextsim2026,convapparel2026}.
The most recent wave deploys agents directly to simulate controlled
experiments: AgentA/B~\cite{igamberdiev2025agentab} sends interactive
agents to browse two website variants;
PAARS~\cite{wang2025paars} conditions agents on structured personas;
S-Researcher~\cite{sresearcher2026} scales to $10^5$ concurrent
agents.
None of these systems reports an explicit decomposition of estimation
error into independently controllable components, making it difficult
to diagnose or reduce error systematically.
A complementary line of work uses AI predictions not to replace
experiments but to reduce variance within them:
\citet{arbour2026ai} show that including AI-generated predictions as
covariates in standard regression adjustment yields efficiency gains
with a ``do no harm'' guarantee.
Our work takes the more ambitious step of simulating entire
experiments, but connects to this literature through the calibration
protocol (\cref{sec:improvements-calibration}).

We propose a framework for agentic A/B prediction, formalized as a
Simulated Randomized Controlled Trial (S-RCT), and validate it
against real historical outcomes.
We develop a two-layer error decomposition that separates agent
approximation error from subsampling error, and instantiate the
framework with pre-period calibration and within-subject estimation.
We validate on a benchmark of \ResultNExperiments{} historical
marketing A/B tests.
Our results surface both positive findings (calibration and
within-subject estimation compress magnitude error substantially)
and fundamental bottlenecks, including systematic agent
over-responsiveness and the limits of persona completeness.
We emphasize lessons learned and failure modes, which may be
as informative for the field as the accuracy gains.

\paragraph{Contributions.}
(1)~We formalize agentic A/B prediction as an S-RCT and derive a
two-layer error decomposition separating approximation error from
subsampling error (\cref{sec:method}).
(2)~We validate the framework on \ResultNExperiments{} historical A/B
tests, documenting accuracy gains and fundamental bottlenecks
(\cref{sec:experiments}).
(3)~We propose calibration and within-subject techniques that
address each error layer independently
(\cref{sec:improvements}).


\section{Method}
\label{sec:method}

\subsection{Framework and formalism}
\label{sec:method-framework}

Consider a universe of $M$ possible users eligible for an experiment.
An experiment-specific triggering mechanism determines which users
are exposed to the intervention, defining a triggered set
$\mathcal{P} \subset \{1,\ldots,M\}$ with
$|\mathcal{P}| = N_{\mathrm{exp}} \le M$.
Each triggered customer $j \in \mathcal{P}$ is randomly assigned to a
treatment arm $T$ or $C$ and is characterized by two unit-level quantities:
\begin{itemize}[nosep]
  \item a \emph{persona} $\Pi_j \in \mathcal{X}$---a vector of
  observable profile features whose value is independent of
  treatment assignment;
  \item an \emph{outcome} $Y_j^t \in \mathbb{R}$, $t \in \{C, T\}$,
  for a metric of interest.
\end{itemize}
The quantity of interest is the average treatment effect (ATE), a
population-level causal parameter that is a latent state of nature:
it can be estimated from data but never directly
observed~\citep{neyman1923application,rubin1978bayesian,imbens2015causal}.
Formally,
\begin{equation}\label{eq:ate}
  \ATE \;:=\;
  \mathbb{E}_{\Pi \sim Q}\!\bigl[
    \mu^T(\Pi) - \mu^C(\Pi)
  \bigr],
\end{equation}
where $\mu^t(\Pi) := \mathbb{E}[Y^t \mid \Pi]$ is the conditional
mean outcome under arm $t \in \{T, C\}$ (treatment or control) and
$Q$ is a distribution over personas.
The choice of $Q$ determines which population the ATE summarizes:
setting $Q$ to the empirical distribution of the triggered set
$\mathcal{P}$ yields the finite-population ATE (the effect for the
users who actually entered the experiment); treating $\mathcal{P}$ as
a sample from a broader superpopulation yields the superpopulation
ATE (the effect one would expect for future
users)~\citep{imbens2015causal}.

A \emph{Simulated Randomized Controlled Trial} (S-RCT) replaces live
traffic with computational surrogates: given a description of
\emph{who} the user is (persona $\Pi_j$), \emph{what} the user
experiences (context), and \emph{what decision} the user must make
(task), a simulator produces a simulated outcome
$\widetilde{Y}_j$.
\cref{tab:srct_example} illustrates these inputs for a marketing
creative test.
The simulator is characterized by a parameter $\theta$ encoding an
approximating response distribution
$\widetilde{P}_\theta(\cdot \mid \Pi_j, t)$.

\begin{table}[t]
\centering
\small
\caption{Example S-RCT inputs for a marketing creative A/B test
measuring click-through rate on a product page.}
\label{tab:srct_example}
\begin{tabular}{@{}ll@{}}
\toprule
\textbf{Component} & \textbf{Concrete example} \\
\midrule
Persona $\Pi_j$ & 35-year-old frequent shopper, electronics \\
                & enthusiast, high brand affinity \\
Context (control) & Current banner: ``Free shipping on orders \$25+'' \\
Context (treatment) & New banner: ``Members save 20\% today'' \\
Task & Does this user click the banner? (yes/no) \\
Outcome $\widetilde{Y}_j$ & $1$ (click) or $0$ (no click) \\
\bottomrule
\end{tabular}
\end{table}
The S-RCT estimator in its simplest form is the difference in
simulated means:
\begin{equation}\label{eq:ate_srct}
  \agentHatATE
  \;=\; \bar{\widetilde{Y}}_T - \bar{\widetilde{Y}}_C
  \;=\;
  \frac{1}{|\mathcal{S}_T|}
  \sum_{j \in \mathcal{S}_T} \widetilde{Y}_j
  \;-\;
  \frac{1}{|\mathcal{S}_C|}
  \sum_{j \in \mathcal{S}_C} \widetilde{Y}_j,
\end{equation}
computed over $N_{\mathrm{agt}} := |\mathcal{S}_T| + |\mathcal{S}_C|$
simulated \emph{agents}---computational units that map (persona,
context, task) triples to decisions.
Our framework is engine-agnostic: the simulator could be a rule-based
model, a learned behavioral surrogate, or a foundation model.

\paragraph{Persona completeness.}
S-RCTs introduce a representation bottleneck: the simulator receives
only the persona $\Pi_j$ as its window into who user $j$ is.
Recall that $\mathcal{X}$ is the space of all observable profile
features available to the simulator.
If the true response depends on factors not captured in
$\mathcal{X}$ (e.g., the user's current intent, mood, or
external context), even a perfect simulator incurs an irreducible
gap.
\begin{assumption}[Persona completeness]\label{assp:completeness}
$Y_j^t \;\perp\!\!\!\perp\; Z_j \;\big|\; (\Pi_j, t)$
for all $Z_j \notin \mathcal{X}$.
\end{assumption}
\cref{assp:completeness} is strong, and likely violated in practice.
Violations contribute to the approximation error formalized next.

\paragraph{Two-layer error decomposition.}
Define the simulator-population ATE as the treatment effect the
simulator would produce if every triggered customer were evaluated:
\begin{equation}\label{eq:ate_theta}
  \agentATE_\theta \;:=\;
  \mathbb{E}_{\widetilde{P}_\theta}\!\Bigl[
    \frac{1}{N_{\mathrm{exp}}}
    \sum_{j \in \mathcal{P}}
    \bigl(\widetilde{Y}_j^T - \widetilde{Y}_j^C\bigr)
  \Bigr].
\end{equation}
The S-RCT estimator's total error admits:
\begin{equation}
\label{eq:decomp}
\agentHatATE - \ATE \;=\;
\underbrace{\bigl(\agentATE_\theta - \ATE\bigr)}_{\text{approximation error}}
\;+\;
\underbrace{\bigl(\agentHatATE -
  \agentATE_\theta\bigr)}_{\text{subsampling error}}.
\end{equation}
The approximation error is the gap between the simulator's
population-level prediction and the true ATE---a function of the
behavioral model, the persona rendering, and conditioning.
The subsampling error is the finite-sample gap between the realized
$N_{\mathrm{agt}}$-agent estimate and the simulator-population ATE.
These two layers are independently addressable: calibration targets
the first; principled agent selection targets the second
(\cref{sec:improvements}).

\subsection{Instantiation}
\label{sec:method-instantiation}

We use an internal simulation platform as the engine, with
three design choices.
First, the simulator is a general-purpose foundation model prompted
with the customer's behavioral profile; we use this as a practical
bootstrapping choice, though the engine-agnostic framework readily
accommodates domain-specific models fine-tuned on behavioral data.
Second, agents are constructed via one-to-one \emph{agentic-twin}
pairing: every agent is bound to a specific real customer who
participated in the historical experiment (trigger-based sampling).
This yields individual-level outcome pairs
$\{(Y_j, \widetilde{Y}_j)\}$ supporting per-customer error analysis.
Third, because the simulator is a stateless function, each agent can
be queried under \emph{both} arms---and repeatedly---yielding paired
individual treatment effects
$\widetilde{\Delta}_j = \widetilde{Y}_j^T - \widetilde{Y}_j^C$
with no analogue in real experiments.
This full counterfactual access is exploited in
\cref{sec:improvements-within-subject}.


\section{Baseline Results}
\label{sec:experiments}

We apply the S-RCT framework to a benchmark of $I=\ResultNExperiments$
historical marketing A/B test treatment pairs run on a large
e-commerce service.
Each experiment is a marketing creative test measuring click-through
rate (CTR) on a high-traffic product surface.
The benchmark includes a variety of creative treatments spanning
text, imagery, and layout variations.
Historical percent impacts span a range typical of marketing creative
tests, with most effects small in magnitude.
The benchmark contains a roughly even split of launch, harmful, and
inconclusive historical decisions.

Each simulation uses $N_{\mathrm{agt}} = 1{,}000$ agents in a
within-subject design where every agent is exposed to both treatment
and control, with each agent mapping one-to-one to a real customer
who participated in the historical experiment.
We use a general-purpose foundation model with no hyperparameter
tuning; agents are chosen uniformly at random from the triggered
population.

\begin{table}[t]
\centering
\small
\caption{Evaluation metrics. $\hat\sigma_i$: standard error of
$\hatATE_i$; $p_i, q_i$: posterior probabilities of a positive
effect; $\mathrm{dec}(x) = -1$ if $x < 0.33$, $1$ if $x > 0.66$,
$0$ otherwise.}
\label{tab:metrics}
\begin{tabular}{@{}l@{\;\;}l@{\qquad}l@{\;\;}l@{\qquad}l@{\;\;}l@{}}
\toprule
\multicolumn{2}{c}{Corrected MSE} &
\multicolumn{2}{c}{Sign overlap} &
\multicolumn{2}{c}{Launch alignment} \\
\midrule
\multicolumn{2}{c}{$I^{-1}\sum_i \bigl[(\agentHatATE_i - \hatATE_i)^2
    - \hat\sigma_i^2\bigr]$} &
\multicolumn{2}{c}{$I^{-1}\sum_i (1 - |p_i - q_i|)$} &
\multicolumn{2}{c}{$I^{-1}\sum_i \mathbf{1}\{\mathrm{dec}(p_i) =
    \mathrm{dec}(q_i)\}$} \\
\bottomrule
\end{tabular}
\end{table}

\paragraph{Results.}
\cref{tab:baseline} reports accuracy metrics across all
\ResultNExperiments{} experiment-treatment pairs.
Agentic impact estimates are systematically larger in magnitude than
historical ATEs (MAE~$= \ResultMAE$, several times the median
historical percent impact), and launch alignment is
$\ResultLaunchAlign$---near the random floor of $0.33$---driven by the
magnitude gap pushing agentic posterior probabilities to extremes.
On directional metrics the simulator reaches sign accuracy
$\ResultSignAccuracy$ and sign overlap $\ResultSignOverlap$.

These directional numbers should be read with care: if the goal is to
quantify whether the forecast carries information about the true
effect, sign accuracy and sign overlap can both be poor proxies, for at
least two reasons.

First, the ground truth is noisy, so we bound the range each metric can
occupy. For sign accuracy, a natural upper bound is the sign-replication
rate---the probability that two independent repetitions of the same
experiment agree in sign, which is $\ResultSignReplication$ here and is
an upper bound for any forecast no more informative than an independent
rerun; from below, uninformed rules already reach
$\ResultNaiveCoinFlipSignAcc$ (a coin flip) and
$\ResultNaiveAlwaysPosSignAcc$ (always predict positive). For sign
overlap, uninformed forecasts bound it from below: a forecast drawing
$q_i$ at random scores $\ResultNaiveUnifDrawSignOv$, and one that always
answers $0.5$ scores $\ResultNaiveCoinFlipSignOv$. Against these ranges,
sign accuracy $\ResultSignAccuracy$ sits near the top of a narrow band
($\ResultNaiveAlwaysPosSignAcc$--$\ResultSignReplication$) and sign
overlap $\ResultSignOverlap$ falls \emph{below} the
$\ResultNaiveCoinFlipSignOv$ a no-conviction forecast attains, so a high
score carries limited information about forecast quality.

Second, sign metrics do not separate genuine predictive information from
a systematic directional lean: our simulator tends to over-predict
positive effects, which inflates agreement with the mostly-positive
historical signs without implying test-level information.

In summary, the baseline simulator is not yet useful for launch
decisions without calibration, and its directional agreement should be
read against these baselines rather than as evidence of signal,
motivating the improvements in \cref{sec:improvements}.

\begin{table}[t]
\centering
\small
\caption{Baseline accuracy (within-subject design,
$N_{\mathrm{agt}} = 1{,}000$, no calibration).
Uninformed floors by metric: sign accuracy $\ResultNaiveCoinFlipSignAcc$
(coin flip) to $\ResultNaiveAlwaysPosSignAcc$ (always-positive); sign
overlap has no single floor ($\ResultNaiveUnifDrawSignOv$--$\ResultNaiveCoinFlipSignOv$,
depending on the $p_i$ distribution); launch alignment $0.33$.}
\label{tab:baseline}
\begin{tabular}{@{}lcc@{}}
\toprule
Metric & Value & SE \\
\midrule
Corrected MSE & $\ResultCMSE$ & $\ResultCMSESE$ \\
Sign overlap & $\ResultSignOverlap$ & $\ResultSignOverlapSE$ \\
Sign overlap (BC) & $\ResultSignOverlapBC$ & $\ResultSignOverlapBCSE$ \\
Launch alignment & $\ResultLaunchAlign$ & $\ResultLaunchAlignSE$ \\
MAE & $\ResultMAE$ & $\ResultMAESE$ \\
Sign accuracy & $\ResultSignAccuracy$ & $\ResultSignAccuracySE$ \\
\bottomrule
\end{tabular}
\end{table}


\section{Improvements}
\label{sec:improvements}

We address the two error layers in \cref{eq:decomp} with three
improvements: principled subsampling
(\cref{sec:improvements-subsampling}), pre-period calibration
(\cref{sec:improvements-calibration}), and within-subject estimation
(\cref{sec:improvements-within-subject}).

\subsection{Principled subsampling}
\label{sec:improvements-subsampling}

Simulations are cheaper than real experiments, but they are not free.
A large service where many teams run A/B tests in parallel, each
involving large user populations, faces a clear scaling constraint:
simulating every triggered user is infeasible.
Smart sampling of \emph{which} users to simulate is therefore
critical for this approach to scale.

\begin{figure*}[t]
\centering
\includegraphics[width=\linewidth]{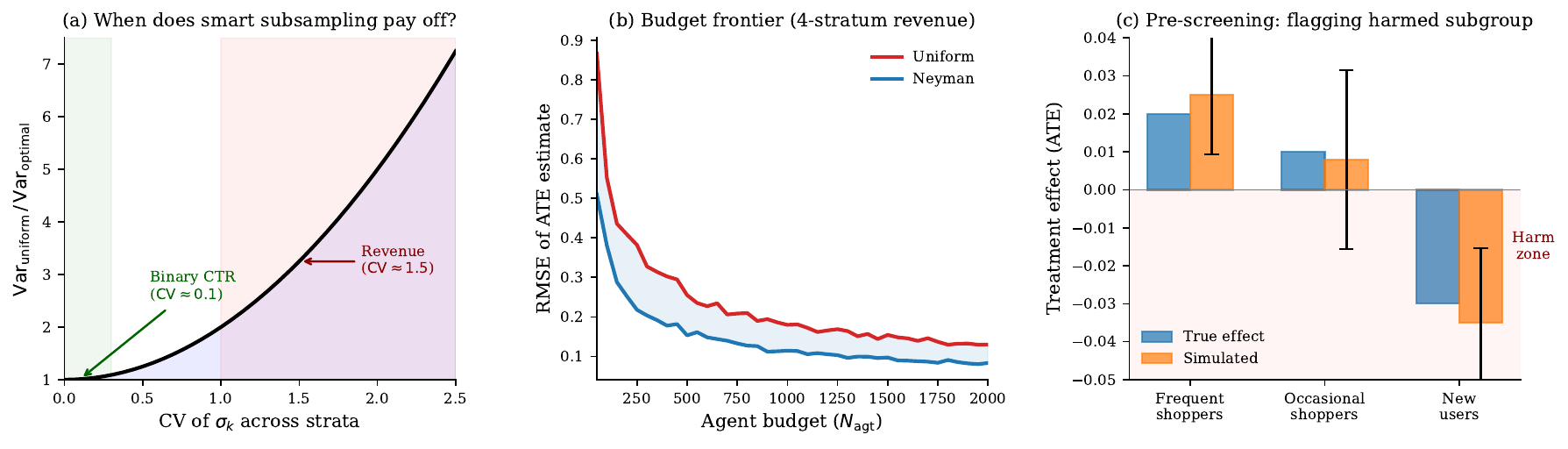}
\caption{Principled subsampling: (a)~efficiency gain from Neyman
allocation grows with stratum variance heterogeneity; (b)~for a
stylized revenue metric, Neyman allocation achieves a given RMSE
with fewer agents; (c)~simulation pre-screens a treatment across
subgroups, flagging the harmed segment before live deployment.}
\label{fig:subsampling}
\end{figure*}

The subsampling error in \cref{eq:decomp} arises precisely because we
simulate only $N_{\mathrm{agt}} \ll N_{\mathrm{exp}}$ agents.
We can always decompose the population response as a mixture of
heterogeneous subgroups:
$P(Y^t) = \sum_k \pi_k \, P(Y^t \mid \text{stratum } k)$, where
$\pi_k$ is the population share of stratum $k$.
This decomposition is useful beyond variance reduction.
Understanding which subpopulation drives the KPI of interest is
valuable in its own right.
Teams lacking deep experimental expertise often struggle to identify
subgroups where a treatment helps or harms.
If the simulation is sufficiently accurate, it offers a cheap way to
pre-screen an intervention across subpopulations, identifying
potentially harmed groups before live
deployment~\citep{wang2025large} and advancing heterogeneous
treatment effect estimation at scale
(\cref{fig:subsampling}c).

For variance reduction specifically, classical survey sampling
results~\citep{imbens2015causal} show that allocating agents
proportionally to $\pi_k \sigma_k$ (where $\sigma_k$ is the
within-stratum standard deviation) minimizes the variance of the
stratified estimator (Neyman
allocation;~\citealp{neyman1923application}):
\begin{equation}\label{eq:neyman_gain}
  \frac{\mathrm{Var}_{\mathrm{uniform}}}
       {\mathrm{Var}_{\mathrm{optimal}}}
  = 1 + \frac{\mathrm{Var}_\pi(\sigma_k)}
             {(\mathbb{E}_\pi[\sigma_k])^2}.
\end{equation}
The practical gain depends on how heterogeneous stratum variances
are (\cref{fig:subsampling}a).
For binary outcomes with low base rates (as in our CTR benchmark),
within-stratum standard deviations are compressed
near $\sqrt{p}$, limiting the gain to a few percent.
For continuous metrics such as revenue, where high-value segments can
have $10$--$100\times$ the variance of low-value ones, the gain
reaches $2$--$5\times$ (\cref{fig:subsampling}b).

The key takeaway is that the ratio
$\mathrm{Var}_\pi(\sigma_k) / (\mathbb{E}_\pi[\sigma_k])^2$ can be
estimated from historical data before any simulation is run, giving
practitioners a principled criterion for when smart subsampling is
worth the added complexity.

We note a tension here: the estimator we employ
(\cref{eq:ate_srct}, difference in simulated means) is model-free and
makes no assumptions about the response surface.
One could in principle do better by imposing more structure, for
instance modeling the conditional response $\mu^t(\Pi)$ directly and
using the simulation budget to refine uncertain regions.
However, stronger modeling assumptions carry higher misspecification
risk. The stratified difference-in-means estimator strikes a
middle ground: it exploits population heterogeneity for efficiency
while remaining valid regardless of the within-stratum response
model.

\subsection{Agent calibration}
\label{sec:improvements-calibration}

The magnitude gap documented in \cref{sec:experiments}, where
agentic ATEs systematically overshoot historical effects, is a
manifestation of the approximation-error term in \cref{eq:decomp}.
To confidently rely on agent-based simulation, the behavioral model
must be calibrated against real activity patterns.
A natural solution is to rely on a foundation model trained
specifically for this task, whether via post-training (e.g.,
supervised fine-tuning on historical behavioral data) or by building
dedicated behavioral models from scratch.
Here we investigate a third, much cheaper (and admittedly less
powerful) alternative: a lightweight two-phase calibration protocol,
related in spirit to prediction-powered
inference~\citep{angelopoulos2023prediction}, that uses pre-exposure
user data as a free supervision signal.

\begin{figure}[t]
\centering
\scalebox{0.78}{%
\begin{tikzpicture}[
  box/.style={draw, rounded corners=3pt, minimum height=9mm,
              font=\small, align=center},
  wide/.style={box, minimum width=32mm},
  narrow/.style={box, minimum width=22mm},
  p1/.style={fill=green!8},
  p2/.style={fill=blue!8},
  bridge/.style={draw, thick, rounded corners=4pt, fill=orange!10,
                 minimum width=40mm, minimum height=14mm,
                 font=\small, align=center},
  arr/.style={-{Latex[length=2.5mm]}, thick},
  lbl/.style={font=\scriptsize\bfseries, text=gray!60!black}
]

\def\rowA{1.2}
\def\rowB{-0.8}

\node[lbl] at (-2.5, \rowA) {Phase 1};
\node[wide, p1] at (0, \rowA) (s1)
  {LLM on $X \setminus \{Y^{\mathrm{pre}}\}$\\[-1pt]
   \scriptsize both arms $\to$ control};
\node[narrow, p1] at (4.2, \rowA) (pred)
  {$\tilde{Y}_j^{\mathrm{pre}}$};

\draw[arr] (s1) -- (pred);

\node[lbl] at (-2.5, \rowB) {Phase 2};
\node[wide, p2] at (0, \rowB) (s2)
  {LLM on full $X$\\[-1pt]
   \scriptsize real $T/C$ assignment};
\node[narrow, p2] at (4.2, \rowB) (raw)
  {$\tilde{Y}_j^t$};
\node[narrow, p2] at (9.2, \rowB) (apply)
  {$\hat{f}(\tilde{Y}_j^t)$};
\node[narrow, fill=yellow!15] at (12.2, \rowB) (out)
  {$\agentHatATE$};

\draw[arr] (s2) -- (raw);
\draw[arr] (raw) -- (apply);
\draw[arr] (apply) -- (out);

\node[bridge] at (9.2, \rowA) (fit)
  {$\displaystyle\hat{f} = \arg\min_f
   \sum_j \ell\!\bigl(f(\tilde{Y}_j^{\mathrm{pre}}),\;
   Y_j^{\mathrm{pre}}\bigr)$};

\draw[arr] (pred) -- (fit);

\draw[arr, red!60] (fit.south) to[out=-60, in=90] (apply.north);

\end{tikzpicture}%
}
\caption{Two-phase calibration pipeline. In Phase 1 we fit the calibration function $\hat{f}$ on pre-period data. In Phase 2 we apply $\hat{f}$
to the raw outputs $\tilde{Y}_j^t$ to produce the calibrated estimate $\agentHatATE$.
}
\label{fig:pipeline}
\end{figure}
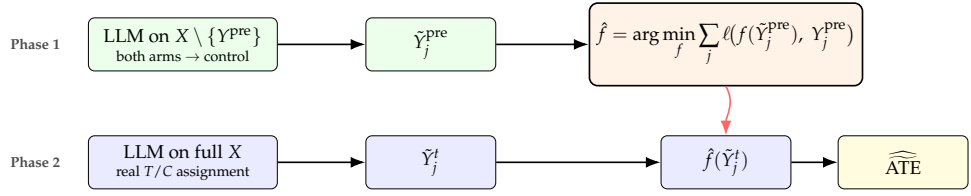

\begin{figure*}[t]
\centering
\includegraphics[width=\linewidth]{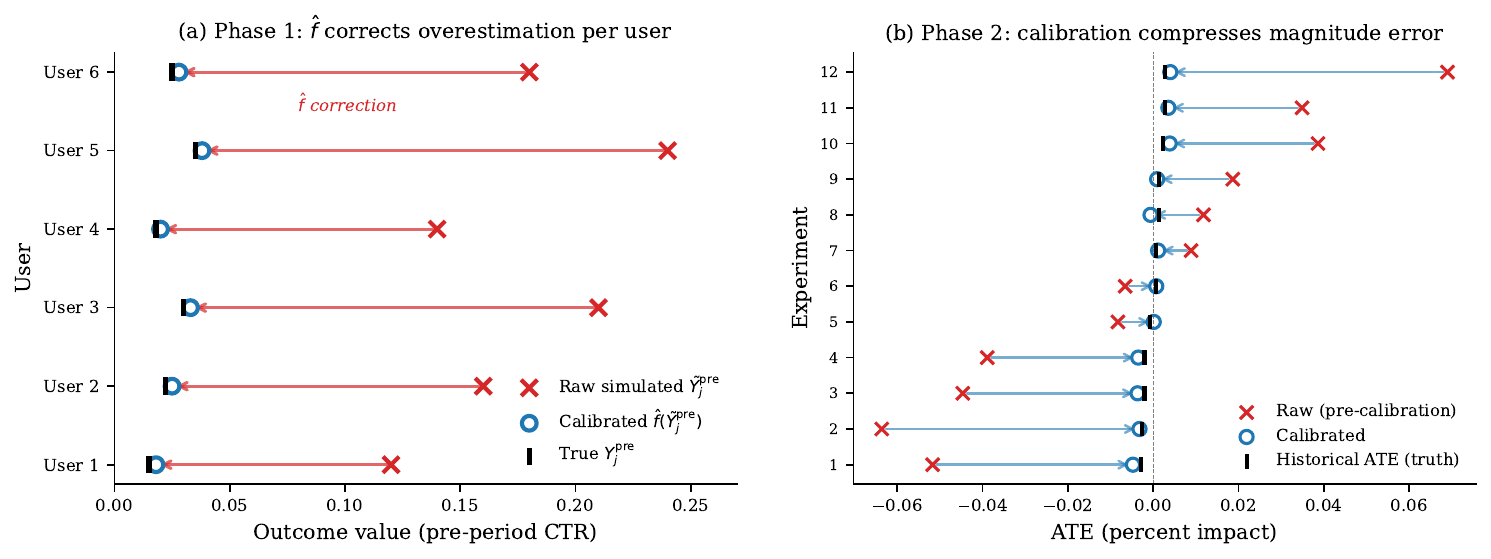}
\caption{Two-phase calibration. (a)~Phase~1: the raw simulator
overestimates user activity; Platt scaling learns a correction from
pre-period data. (b)~Phase~2 result: per-experiment ATE estimates
before (red crosses) and after (blue circles) calibration, relative
to the historical ATE (black bars). Calibration compresses the
magnitude gap by ${\sim}77\times$.}
\label{fig:calibration}
\end{figure*}

The protocol has two phases (\cref{fig:pipeline}).
In \emph{Phase~1} (calibration), we run the simulator on the
pre-period: the pre-period outcome $Y_j^{\mathrm{pre}}$ is held out
of the feature set, and both arms render the control context (an A/A
simulation with zero treatment effect by construction).
The simulator produces $\tilde{Y}_j^{\mathrm{pre}}$, and because the
real $Y_j^{\mathrm{pre}}$ is known, we fit a calibration function:
\begin{equation}\label{eq:cal_fit}
  \hat{f} = \arg\min_f \sum_{j} \ell\!\bigl(
    f(\tilde{Y}_j^{\mathrm{pre}}),\; Y_j^{\mathrm{pre}}
  \bigr),
\end{equation}
where $\ell$ is a suitable loss (e.g., log-loss for Platt scaling).
\cref{fig:calibration}a illustrates this step: the raw simulator
systematically overestimates user activity relative to real
pre-period outcomes, and Platt scaling learns a monotone correction.

In \emph{Phase~2} (prediction), we run the simulator with full
features and real treatment assignment.
Raw simulated outcomes are passed through $\hat{f}$ to produce
calibrated estimates.
The calibration function $\hat{f}$ is fit independently per
experiment on Phase-1 data and applied to Phase-2 data; the
train/test split is temporal (pre-period vs.\ treatment period).

On 16 experiments from the benchmark using Platt scaling, calibration
compresses the squared prediction error by ${\sim}77\times$, from
systematic magnitude overshoot to estimates near the historical noise
floor (\cref{fig:calibration}b).

\subsection{Within-subject estimates}
\label{sec:improvements-within-subject}

With agents split across arms, a between-subject design is
susceptible to random composition imbalances.
For subtle effects---where the true ATE is small---the estimated
difference can be dominated by chance differences in who was assigned
to each arm rather than by the treatment itself.
We address this with a within-subject design where every agent is
exposed to \emph{both} treatment and control.
This eliminates between-agent composition differences entirely: each
agent serves as its own control, and the estimator becomes the
within-subject paired difference.
The paired structure removes the persona-level variance component
while targeting the same population ATE.
To address ordering effects, we randomize the order in which
treatment and control are presented to each agent.

The within-subject design's primary advantage is variance reduction:
standard errors shrink by ${\sim}2.4\times$ on average.
Sign accuracy rises from $65\%$ (between-subject) to
$\ResultSignAccuracy$ (within-subject), consistent with the
elimination of random arm-composition imbalance.
Launch alignment improves from $0.33$ to $\ResultLaunchAlign$, though
the magnitude gap continues to push posterior probabilities toward
extremes.
Combining within-subject estimates with calibration
(\cref{sec:improvements-calibration}) is the natural next step.


\section{Discussion and Conclusion}
\label{sec:conclusion}

This paper presents a framework for agentic A/B prediction formalized
as Simulated Randomized Controlled Trials (S-RCTs) with a two-layer
error decomposition separating behavioral approximation error from
subsampling error.
Validation on \ResultNExperiments{} historical marketing A/B tests
reveals directional agreement that stays within the range of uninformed
baselines on this noisy set, and a systematic magnitude gap where
agentic estimates overshoot historical effects.
Within-subject estimation improves sign accuracy; calibration
addresses magnitude error; principled subsampling targets the
remaining variance.

\paragraph{Behavioral implications.}
The systematic over-responsiveness of agent behavior---predicting
larger effects than observed in real customers---suggests that
current general-purpose foundation models exhibit a form of
\emph{behavioral amplification}: when conditioned on a persona and
asked to choose, they respond more decisively to treatment
differences than real customers do.
Understanding and correcting this behavioral gap is, we believe, a
core challenge for the agent behavior community: it reflects not just
miscalibration but a fundamental mismatch between how these models
represent decision-making and how humans actually behave in low-stakes
product contexts.
Bridging this gap may require behavioral fine-tuning on revealed
preference data, structured constraints on agent response
distributions, or hybrid architectures that combine language model
reasoning with learned behavioral priors.

\paragraph{Applications beyond ATE estimation.}
Estimating the average treatment effect is arguably the hardest target
for agentic simulation: it requires both correct direction and
accurate magnitude.
However, the framework produces several secondary outputs that are useful
even when magnitude accuracy is unreliable.
\emph{Directional screening} (flagging likely losers before committing
live traffic) requires only correct sign rather than accurate magnitude,
a lower bar than ATE estimation. On the present benchmark set,
however, sign accuracy and sign overlap are only partially informative:
as discussed above, benchmark noise limits how much they reveal about
whether the simulator's directional predictions carry real signal.
Progress on this application will therefore require both larger, less
noisy benchmarks and better evaluation metrics---itself a direction for
future work.
\emph{Reasoning traces} (qualitative explanations of why a simulated
agent chose one option over another) offer experimenters a novel
interpretive lens with no analogue in classical A/B testing.
\emph{Flipper analysis} (identifying which agent personas flip their
decision between treatment and control) surfaces treatment-effect
heterogeneity and can guide segment-level analysis in real
experiments.
\emph{Agentic priors for early decisions}: even a noisy directional
signal, when combined with early real-experiment data, can accelerate
go/no-go decisions by serving as an informative prior in a Bayesian
framework.
These applications lower the accuracy bar required for practical
value and represent the nearest-term path from the current system to
practical deployment.

\paragraph{Limitations.}
Near-term limitations include magnitude overshoot even with correct
sign; a single-domain benchmark (marketing CTR); potential
correlation in agent responses from a shared model that could
understate variance; and retrospective-only evaluation against
historical targets.
The benchmark evaluates a prediction against a known historical
outcome---in a prospective setting, the triggering population is
unknown and no ground truth is available for calibration.
Predicting the triggering population is the main missing piece for
deployment; it is best understood as a thin upstream layer that can be
developed independently of the behavioral simulator.

\paragraph{Broader considerations.}
If the framework is used to pre-screen which experiments to run,
treatments benefiting customer segments poorly captured by the
behavioral model---low-frequency users, underrepresented
demographics---may be filtered out before reaching real measurement.
The two-phase calibration protocol partially mitigates this by
surfacing per-segment bias on the pre-period, but practitioners must
remain attentive to whose behavior the simulator represents
faithfully and whose it does not.
The goal is not to replace real experiments but to make the
experimentation pipeline faster and more informed, so that customers
benefit from product improvements sooner.

\bibliography{references}
\bibliographystyle{colm2026_conference}

\appendix

\section{Evaluation Metrics: Detailed Definitions}
\label{app:metrics}

\paragraph{Corrected MSE.}
The noise-corrected MSE removes the irreducible historical sampling
variance from the squared error:
\[
  \text{CMSE} = \frac{1}{I}\sum_{i=1}^{I}
  \bigl[(\agentHatATE_i - \hatATE_i)^2 - \hat\sigma_i^2\bigr],
\]
where $\hat\sigma_i$ is the standard error of the historical ATE
estimate for experiment $i$.
Without this correction, MSE would be dominated by the historical
noise floor rather than the simulator's error.

\paragraph{Sign overlap.}
Sign overlap measures the agreement between historical and agentic
posterior probabilities of a positive effect:
\[
  \text{SO} = \frac{1}{I}\sum_{i=1}^{I}(1 - |p_i - q_i|),
\]
where $p_i$ and $q_i$ are the posterior probabilities of a positive
effect under the historical and agentic estimates, respectively.
A value of $1.0$ indicates perfect calibration of directional
confidence. There is no single random floor: for an uninformative agent
drawing $q_i \sim \mathrm{Uniform}[0,1]$, the expected per-experiment
score is $\tfrac{1}{2} + p_i(1-p_i)$, which equals $0.50$ only as
$p_i \to 0$ or $1$ and rises to $0.75$ at $p_i = 0.5$. Averaged over our
$p_i$ this floor is $\ResultNaiveUnifDrawSignOv$, and a constant
$q_i = 0.5$ agent scores $\ResultNaiveCoinFlipSignOv$; because our
posteriors cluster near $0.5$, these uninformed baselines sit close to
the achievable range, so a high sign overlap is not on its own evidence
of directional quality.

\paragraph{Launch alignment.}
Launch alignment discretizes the posterior into three decisions
(harmful: $p < 0.33$; inconclusive: $0.33 \leq p \leq 0.66$;
launch: $p > 0.66$) and measures agreement:
\[
  \text{LA} = \frac{1}{I}\sum_{i=1}^{I}
  \mathbf{1}\{\mathrm{dec}(p_i) = \mathrm{dec}(q_i)\}.
\]
The random floor is $0.33$ (three equally likely categories).

\section{Benchmark Details}
\label{app:benchmark_data}

The benchmark comprises \ResultNExperiments{} treatment-control pairs
from marketing creative tests on a high-traffic e-commerce product
surface.
All experiments measure click-through rate (CTR) as the primary
metric.
Historical percent impacts are small, consistent with the subtle
nature of marketing creative changes.
The distribution of historical decisions is approximately balanced
across harmful, inconclusive, and launch outcomes.
Each experiment draws from a large triggered population,
representing a broad cross-section of customer segments.

\end{document}